\documentclass[nohyperref]{article}

\usepackage{microtype}
\usepackage{graphicx}
\usepackage{subfigure}
\usepackage{booktabs} 

\usepackage{hyperref}

\usepackage[accepted]{icml2022}

\usepackage{amsmath}
\usepackage{amssymb}
\usepackage{mathtools}
\usepackage{amsthm}

\usepackage[capitalize,noabbrev]{cleveref}

\theoremstyle{plain}

\theoremstyle{definition}

\theoremstyle{remark}

\usepackage[textsize=tiny]{todonotes}

\icmltitlerunning{Learnable Non-linear Wavelet Transformation via Normalizing Flow}

\makeatletter
\DeclareRobustCommand\onedot{\futurelet\@let@token\@onedot}
\def\@onedot{\ifx\@let@token.\else.\null\fi\xspace}

\def\eg{\emph{e.g}\onedot} 
\def\ie{\emph{i.e}\onedot}

\def\wrt{w.r.t\onedot} 

\makeatother

\newcommand{\code}[1]{\texttt{#1}}

\usepackage{amssymb,amsmath}
\graphicspath{{figs/}} 
\usepackage{booktabs}
\usepackage{lmodern}
\usepackage{xcolor}
\usepackage{natbib}
\usepackage{xspace}
\usepackage{notoccite}
\usepackage[ruled,vlined,algosection,resetcount]{algorithm2e}

\begin{document}
\twocolumn[
\icmltitle{Learnable Non-linear Wavelet Transformation via Normalizing Flow}



\icmlsetsymbol{equal}{*}

\begin{icmlauthorlist}
\icmlauthor{Shuo-Hui Li}{yyy}
\end{icmlauthorlist}

\icmlaffiliation{yyy}{Department of Physics, The Hong Kong University of Science and Technology
Clear Water Bay, Kowloon, Hong Kong}

\icmlcorrespondingauthor{Shuo-Hui Li}{shuohuili@ust.hk}

\icmlkeywords{Machine Learning, Normalizing Flow, Compression, Discrete Wavelet Decomposition}

\vskip 0.3in
]



\printAffiliationsAndNotice{}  

\begin{abstract}
  Wavelet transformation stands as a cornerstone in modern data analysis and signal processing. Its mathematical essence is an invertible transformation that discerns slow patterns from fast ones in the frequency domain. 
  Such an invertible transformation can be learned by a designed normalizing flow model.
  With a generalized lifting scheme as coupling layers, a factor-out layer resembling the downsampling,
  and parameter sharing at different levels of the model,
  one can train the normalizing flow to filter high-frequency elements at different levels,
  thus extending traditional linear wavelet transformations to learnable non-linear deep learning models.
  In this paper, a way of building such flow is proposed, along with a numerical analysis of the learned transformation.
  Then, we demonstrate the model's ability in image lossless compression, show it can achieve SOTA compression scores while achieving a small model size, substantial generalization ability, and the ability to handle high-dimensional data.
\end{abstract}
\section{Introduction}\label{sec:intro}

The applications of wavelet transformation are pervasive, ranging from image processing~\cite{marr1980theory, canny1986computational, mallat1992singularity, dugad1998new, he1994wavelet}, to data manipulations~\cite{donoho1992wavelet, donoho1994ideal, charrier1999jpeg2000, rabbani2002jpeg2000}, and to numerical analysis~\cite{sweldens1994construction, cruz2001using}. As deep learning advances to domain-specific areas, wavelet transformation also has been used in machine learning tasks. For example, super-resolution task~\cite{huang2017wavelet, zhong2018joint, yu2020wavelet}, image generation task~\cite{ardizzone2019guided}, image segmentation task~\cite{figueiredo2005bayesian}, and so on.
Powerful as they are, wavelet transformations have several unfavorable traits: they are linear transformations, which limits their expressive ability;
and they are derived using mathematical constraints, which prevents them to be tailored to specific tasks on specific datasets.

Normalizing flows are a family of non-linear invertible generative models~\cite{dinh2015nice, dinh16_densit_estim_using_real_nvp, kingma18_glow, chen18_neural_ordin_differ_equat, neuralRG, neuralCT}. Their applications also cover many topics, including image generation~\cite{dinh16_densit_estim_using_real_nvp, kingma18_glow}, independent component analysis (ICA)~\cite{dinh2015nice, sorrenson2020disentanglement}, variational inference~\cite{kingma2016improved, rezende2015variational},
Monte Carlo sampling~\cite{song2017nice},
scientific applications~\cite{neuralRG, noe2019boltzmann, neuralCT, Hu_2020}, and so on.
Normalizing flow is a promising candidate for learnable wavelet transformation. They share many common features with wavelet transformation. The two are both invertible transformations, and normalizing flows have a proven ability to discern frequency patterns~\cite{dinh2015nice, neuralRG, neuralCT, sorrenson2020disentanglement}. Additionally, some types of normalizing flow have a factor-out mechanism~\cite{neuralRG, dinh16_densit_estim_using_real_nvp}, which resembles downsampling in wavelet transformation.

\begin{figure*}[t]
\begin{center}
\centerline{\includegraphics[width=0.7\textwidth]{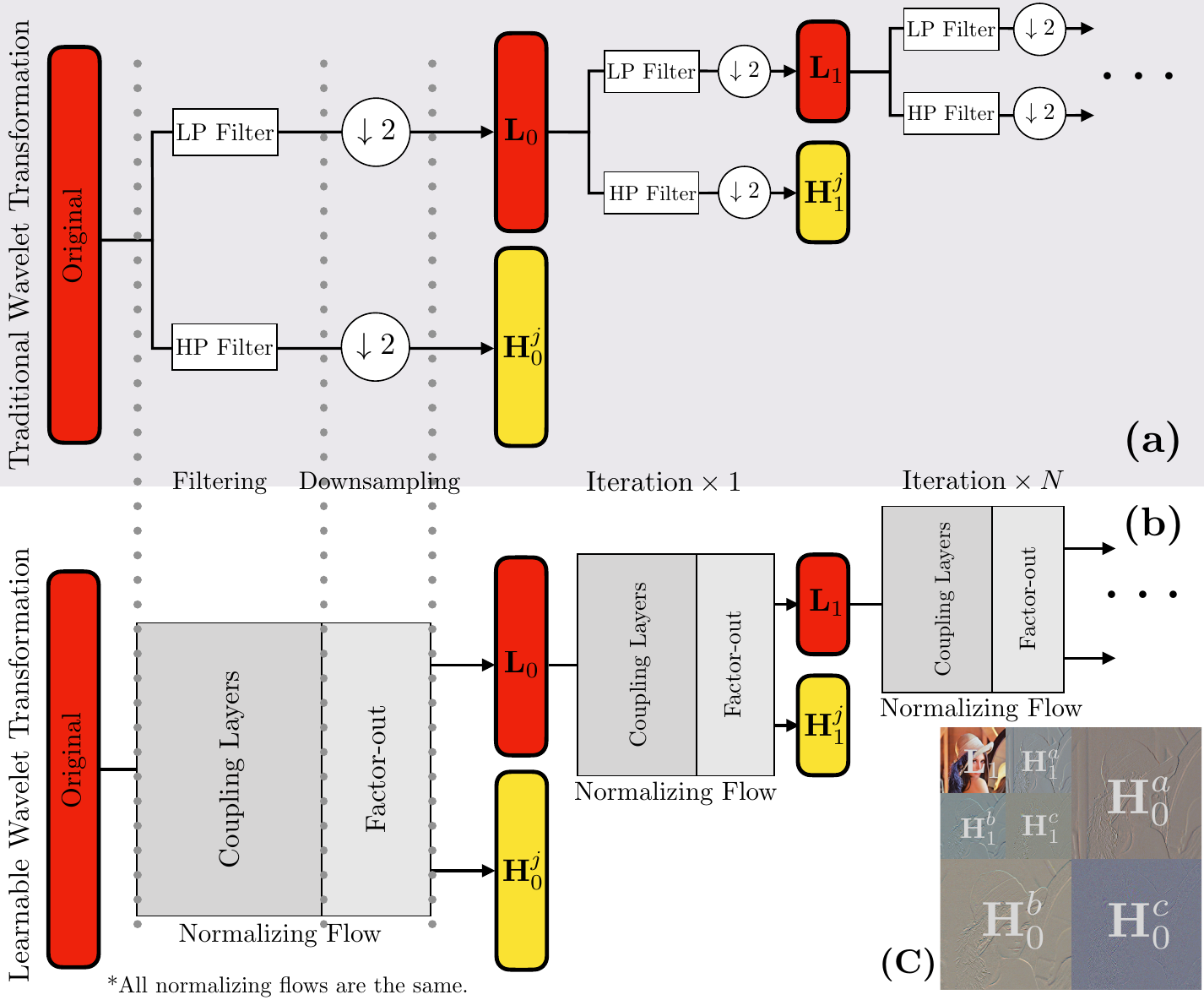}}
\caption{Connections between (a) traditional wavelet transformation process and (b) learnable wavelet transformation architecture.
  (c) shows results from two iterations of Haar wavelet transformation. $\textbf{H}_{n}^j$ is high-frequency parts of $n$-th iteration. For 2D data, $\textbf{H}_{n}^j=\{\textbf{H}_{n}^a,\textbf{H}_{n}^b,\textbf{H}_{n}^c\}$. And $\textbf{L}_n$ is the low-frequency part of $n$-th iteration.}
\label{fig:overall}
\end{center}
\end{figure*}

Building on the inherent connections between wavelet transformation and normalizing flow (Fig.~\ref{fig:overall}), we propose a learnable wavelet transformation model. In this model, we use a generalized lifting scheme as coupling layers to perform filtering, and then factor-out high-frequency patterns from slow ones as downsampling. Following the traditional wavelet, we share parameters of coupling layers at different levels/iterations to repeat the same transformation/filtering. One can train this kind of model as a usual flow model, \ie, by minimizing negative log-likelihood.

As a normalizing flow, the proposed model's wavelet-like behavior gives substantial flexibility: the share of parameters relaxes dimensional constraint, allowing the model to be used on different data dimensions from trained on.
And the share of parameters also means a smaller parameter size and better computation efficiency.
The usage for this learnable wavelet is diverse.  It's a natural fit for tasks traditionally using wavelets, like lossless compression, super-resolution, PDE/ODE solver, and so on. Compare to traditional wavelets, it's trained to fit its task purposes and datasets. And it's a non-linear transformation, which grants it more expressive power in complex tasks.

Later in the paper, we demonstrate this model in the image lossless compression task by following the JPEG-2000 algorithm~\cite{rabbani2002jpeg2000}. We show that this kind of novel compression model outperforms traditional compression algorithms and state-of-the-art (SOTA) deep-learning-based compression models, while smaller than the previous SOTA compression model. Additionally, it has substantial generalization ability, one can use it on images with different resolutions than trained on, without much performance loss. 

This paper is organized as follows. Sec.~\ref{sec:background} gives an introduction of the background. 
Sec.~\ref{sec:nwl} shows the exact scheme of constructing the generalized-lifting-style coupling layers and factor-out for this model, and also shows how to set up prior distributions. 
Sec.\ref{sec:arch} gives details about the architecture used.
Sec.~\ref{sec:analysis} gives a numerical analysis of the learned filters.
Sec.~\ref{sec:experiment} demonstrates the model's performance in lossless compression.
Sec.~\ref{sec:related} gives a brief review of related works.
Sec.~\ref{sec:conclusion} concludes the paper.

\section{Background}\label{sec:background}

This section introduces two fundamental concepts, the normalizing flow and the discrete wavelet transformation.

\subsection{Normalizing flow}\label{sec:flow}

Normalizing flows are a family of parameterized bijective mappings. Except for invertibility, another core feature of these bijective mappings is their traceable probability change. And the maximization of this traceable probability is also the optimization goal. For a bijective mapping $f:\mathcal{X} \rightarrow \mathcal{Z}$, its probability change can be described as
\begin{equation}
  \text{P}_{\mathcal{X}}(x) = \text{P}_{\mathcal{Z}}(z)\left|\rm det\left(\frac{\partial{z}}{\partial{x}}\right)\right|
  \label{eq:flowProbability}
\end{equation}
where $P_{\mathcal{X}}(x)$ and $P_{\mathcal{Z}}(z)$ are the probability of the original sample $x$ and transformed variables $z$, respectively. So, for a transformation to be a flow, 1). it has to be invertible, and 2). its Jacobian determinant should be designed easily computable. Through this probability relation, one can thus maximize the sample probability via maximizing the probability of the transformed variables in a prior distribution $P_{\mathcal{Z}}$. This prior distribution is usually a simple pre-defined distribution for the ease of optimization.

One concrete example of such mapping is the Non-linear Independent Component Estimate (NICE)~\cite{dinh2015nice}. A NICE consists of multiple coupling layers.
One coupling layer of NICE formulates as follows,
\begin{equation}
  [\textbf{z}_A, \textbf{z}_B] = [\textbf{x}_A + t(\textbf{x}_B), \textbf{x}_B]
  \label{eq:vanillaNICE}
\end{equation}
where input $\textbf{x} = [\textbf{x}_A, \textbf{x}_B]$, \ie, we divide input $\textbf{x}$ into two parts. And so does the output $\textbf{z}$.
$t$ is a parameterized function that maps $\textbf{x}_B$ to the space of $\textbf{x}_A$. 
To change all variables, one changes the roles of $\textbf{z}_A$ and $\textbf{z}_B$ at the next coupling.
From Eq.~(\ref{eq:vanillaNICE}), one can see the coupling is invertible: $[\textbf{x}_A, \textbf{x}_B] = [\textbf{z}_A-t(\textbf{z}_B), \textbf{z}_B]$, and the Jacobian determinant is exactly $1$. This also means the NICE is invertible and has a Jacobian determinant of $1$.
The maximization of the probability of $\textbf{z}$ in prior distribution also equally maximizes the probability of $\textbf{x}$, as they equal each other.

In some types of normalizing flows~\cite{dinh16_densit_estim_using_real_nvp, neuralRG}, a mechanism called the factor-out layer is used.
Factor-out layers are usually put after a certain number of coupling layers.  When factor-out, part of the variables are kept from later coupling layers. This lowers data dimensions for later transformations, reducing computation complexity.


\subsection{Discrete wavelet transformation}\label{sec:wavelet}

Discrete wavelet transformations are invertible linear transformations. These transformations are usually formed by two kinds of filters, a low-pass (LP) filter $\textbf{S}$ and a high-pass (HP) filter $\textbf{G}$. We write them as different rows in a single transformation matrix $\textbf{W}$ in $\mathbb{R}^{d\times d}$. Then the transformation can be formulated as follows,
\begin{equation}
  \textbf{z} = \textbf{W}\textbf{x} = \left[\frac{\textbf{S}\textbf{x}}{\textbf{G}\textbf{x}}\right] = \left[\frac{\textbf{L}}{\textbf{H}}\right] 
  \label{eq:1Dwavelet}
\end{equation}
where $\mathbf{z}$ and $\mathbf{x}$ are one-dimensional vectors in $\mathbb{R}^d$. $\textbf{L},~\textbf{H} \in \mathbb{R}^{d/2}$, and they are low-frequency and high-frequency coefficients, respectively.
This linear transformation is followed by a downsampling operation, dividing the output into $\textbf{L}$ and $\textbf{H}$.
The whole linear transformation with downsampling is called one iteration of the wavelet transformation. And we repeat these operations on the low-frequency coefficients $\textbf{L}$ at the next iteration. As shown in Fig.~\ref{fig:overall}(a).

For two-dimensional data, one applies the transformation matrix twice, once for the rows and once for the columns, as in Eq.~(\ref{eq:2Dwavelet}).
And the next iteration is performed on the upper-left  quarter of the matrix (the lowe-frequency part).
\begin{equation}
  \textbf{z} = \textbf{W}\textbf{x}\textbf{W}^T 
  \label{eq:2Dwavelet}
\end{equation}
\section{Learnable wavelet transformation}\label{sec:nwl}

In this section, we propose the learnable wavelet transformation model (LWT). LWT has an architecture as shown in Fig.~\ref{fig:overall}(b). We first present connections between flows and wavelets, from which we define 1D/2D coupling layers and factor-out layers. Then, we define the typical prior distributions of high/low-frequency coefficients for training.

\begin{figure*}[htp]
\begin{center}
\centerline{\includegraphics[width=0.85\textwidth]{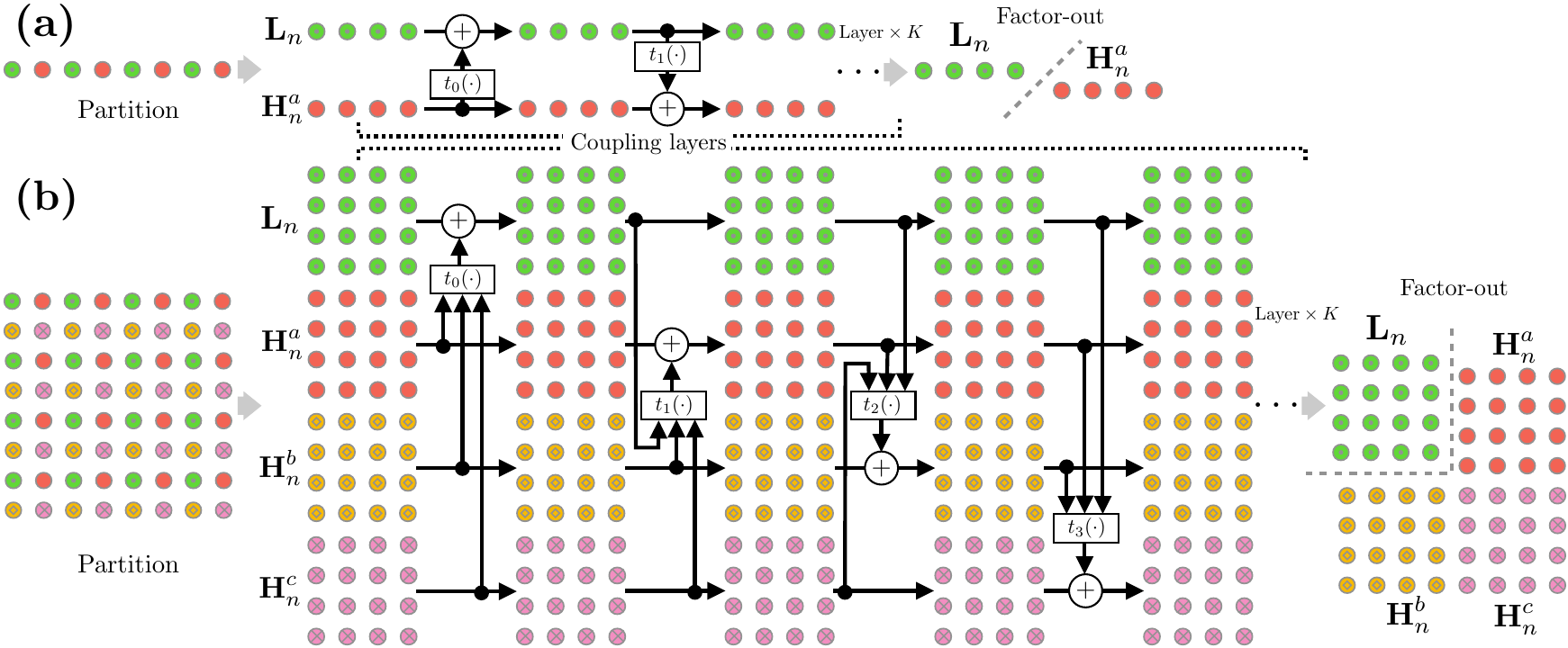}}
\caption{Coupling layer and factor-out in (a) one-dimensional case and (b) two-dimensional case.
  To show the partition scheme, colors and symbols are used to distinguish different locations.
  The data in each circle may have a channel dimension.
  When factor-out, both cases only keep the low-frequency part $\textbf{L}_n$.
  These constitute one iteration of LWT in Fig.~\ref{fig:overall}(b).
}
\label{fig:partition}
\end{center}
\end{figure*}
\subsection{Connections between wavelet transformations and normalizing flows}\label{sec:1d}

One can relate wavelet transformations with the normalizing flows in the following aspects:

\textbf{Generalized lifting scheme}: The invertibility of wavelets is guaranteed by the orthogonal or bi-orthogonal condition on the transformation matrix $\textbf{W}$ of Eq.~(\ref{eq:1Dwavelet})~\cite{van2011discrete}. One can also derive the invertibility through a fast implementation of wavelets: the lifting scheme~\cite{sweldens1998lifting}. The lifting resembles coupling layers of flows. For example, the lifting scheme expresses the Haar wavelet transformation~\cite{harr1910theorie} in the following form
\begin{equation}
  \begin{split}
    [\textbf{y}_o,\textbf{y}_e] &= [\textbf{x}_o -  \textbf{x}_e,\textbf{x}_e]\\
    [\textbf{z}_o,\textbf{z}_e] &= [\textbf{y}_o,\textbf{y}_e + \lfloor \textbf{y}_o / 2 \rceil]
  \end{split}
  \label{eq:lifting}
\end{equation}
The $\textbf{z}_o$ and $\textbf{z}_e$ are high-frequency and low-frequency coefficients, respectively. $\textbf{y}$ are intermediate variables.
And the $\textbf{x}_o$ and $\textbf{x}_e$ are odd-index and even-index variables in the input $\textbf{x}$, respectively:
\begin{equation}
  \textbf{x}_o = [x_1, x_3, x_5, x_7, \cdots]
  \text{,}~~~
  \textbf{x}_e = [x_0, x_2, x_4, x_6, \cdots] 
  \label{eq:partition}
\end{equation}
The similarity of Eq.~(\ref{eq:vanillaNICE}) and Eq.~(\ref{eq:lifting}) allows one to write this transformation as two coupling layers. We put a detailed deduction in Appx.~\ref{app:glm}. And one also sees that any transformations represented by lifting can be represented by NICE coupling with even-odd variable partition.

\textbf{Downsample and factor-out}: Factor-out layers in flow models drop part of variables, leaving the other part of variables entering later transformations. In this sense, downsampling can be viewed as one type of factor-out: after downsampling, variables are split into high-frequency and low-frequency parts, and the next wavelet iteration operates on the low-frequency part, as in Fig.~\ref{fig:overall}(a).

\textbf{Filtering in probability view}: After filtering, high-frequency coefficients contain mostly local information, such as noises and edges, while low-frequency coefficients have non-local information~\cite{van2011discrete}. As flow models are probability models, we interpret filtering in a probability view. These high-frequency coefficients are local and thus independently distributed. And low-frequency coefficients are correlated, changing correlatively at certain length levels. It is these long-range correlations that contain non-local information.
Prior distributions can thus be set for the proposed model for it to have filtering effect. To be exact, we set the prior for the factored-out/downsampled variables to be uncorrelated and independent for them to be high-frequency.
 And the other remaining variables are automatically correlated and low-frequency:
 One sees this in the inverse direction of the flow(right to left in Fig.~\ref{fig:overall}(b)), start with uncorrelated variables, the inverse transformation creates correlations at different length levels.
 Then, the coupling layer inside flow can be viewed as an LP/HP filter pair.
 Additionally, the traceable probabilities in flow model (Eq.~(\ref{eq:flowProbability})) allow the simultaneous maximization of coupling's filtering effect and the original sample probability.



\subsection{One-dimensional coupling and factor-out}

Building on these connections, we define a coupling layer and factor-out for LWT in one-dimensional case.

For a one-dimensional vector data, we divide it into two parts following the convention of lifting scheme: 
the even-index variables are labeled $\textbf{L}_n$ and the odd-index variables are labeled $\textbf{H}^{a}_n$, as in Fig.~\ref{fig:partition}(a).
Then transformation of these two parts can be parameterized by a coupling layer similar to NICE's coupling (the Eq.~(\ref{eq:vanillaNICE})):
\begin{equation}
  [\textbf{L}_n, \textbf{H}_{n}^a]_{k+1} = [\textbf{L}_n + t_k(\textbf{H}_{n}^a), \textbf{H}_{n}^a]_{k}
  \label{eq:1Dcoupling}
\end{equation}
Here, $[\cdot]_k$ means the inner variables are results of $k$-th coupling layer, and subscript $n$ means these variables are from $n$-th iteration. One iteration contains multiple coupling layers.
$t_k(\cdot)$ is a neural-network-parameterized function. After Eq.~(\ref{eq:1Dcoupling}), as in NICE, one can change $\textbf{H}_{n}^a$ using $\textbf{L}_n$ at the next coupling layer to change all variables (Fig.~\ref{fig:partition}(a)).

After all $K$ coupling layers, the resulted $\textbf{H}_{n}^a$ is viewed as the high-frequency part of this $n$-th iteration, and is factored-out. The $\textbf{L}_n$ is viewed as the low-frequency part, and enters the next iteration.
This is the factor-out scheme for 1D LWT,
as in Fig.~\ref{fig:partition}(a).

This partition, transformation by coupling layers, and factor-out process constitute one iteration of the one-dimensional LWT. We perform multiple iterations while sharing the same coupling parameters to repeat the same LP/HP filter pair (Fig.~\ref{fig:overall}(b)).

This scheme can be initialized to perform certain traditional wavelets at the start, \eg, Appx.~\ref{app:init} demonstrates a way of initializing the coupling layers to be the LeGall wavelet~\cite{le1988sub}.


\subsection{Two-dimensional coupling and factor-out}\label{sec:2d}

With the one-dimensional coupling layer defined above, one way to define two-dimensional coupling is following the conventional 2D wavelet, as in Eq.~(\ref{eq:2Dwavelet}). We perform the one-dimensional coupling twice, once for the batch of rows, and then perform on the columns. And the low-frequency part is the upper-left quarter of the output matrix. For the pseudocode of this scheme, see Algo.~\ref{alg:2dwaveletScheme1}.

Here, we propose another more flexible way of defining 2D coupling. We first separate the input matrix into four parts. As diagramed in Fig.~\ref{fig:partition}(b), in every $2 \times 2$ non-overlapping sub-matrix from the input $d \times d$ matrix, we label the upper left, upper right, lower left, and lower right variables as $\textbf{L}_n$, $\textbf{H}_n^a$, $\textbf{H}_n^b$, and $\textbf{H}_n^c$, respectively.
The resulting $\textbf{L}_n$ matrix is a $\frac{d}{2} \times \frac{d}{2}$ matrix, and so does $\textbf{H}_n^a$, $\textbf{H}_n^b$, $\textbf{H}_n^c$. Then the following coupling layer is performed to update matrix $\textbf{L}_n$,
\begin{equation}
    \left[
      \begin{matrix}
        \textbf{L}_n & \textbf{H}_{n}^a \\
        \textbf{H}_{n}^b & \textbf{H}_{n}^c
      \end{matrix}
    \right]_{k+1}
    =
    \left[
      \begin{matrix}
        \textbf{L}_n + t_k([\textbf{H}_{n}^a,  \textbf{H}_{n}^b,  \textbf{H}_{n}^c])& \textbf{H}_{n}^a \\
        \textbf{H}_{n}^b & \textbf{H}_{n}^c
      \end{matrix}
    \right]_{k}
  \label{eq:2dNICE}
\end{equation}
Here, $[\cdot]_k$ means the inner variables are results of $k$-th coupling layer, and subscript $n$ means these variables are from $n$-th iteration. One iteration contains multiple coupling layers. $t_k(\cdot)$ is a parameterized function that maps $[\textbf{H}_n^a, \textbf{H}_n^b, \textbf{H}_n^c]$ into the space of $\textbf{L}_n$.
Three more coupling layers like Eq.~(\ref{eq:2dNICE}) are performed to update $\textbf{H}_{n}^a$, $\textbf{H}_{n}^b$, and $\textbf{H}_{n}^c$ in turn, as in Fig.~\ref{fig:partition}(b).

For factor-out, we keep the low-frequency part $\textbf{L}_n$ for the next iteration. The rest matrices $\textbf{H}_{n}^a$, $\textbf{H}_{n}^b$, and $\textbf{H}_{n}^c$ are considered the high-frequency parts of $n$-th iteration, and are factored-out. One iteration of 2D LWT contains the partition, the $K$ coupling layers, and the factor-out layer.
We repeat the iteration multiple times with the same parameters for coupling, thus repeating the same filter.
A pseudocode of the above process is in Algo.~\ref{alg:2dwaveletScheme2}.

\begin{figure*}[htp!]
\begin{center}
\centerline{\includegraphics[width=0.75\textwidth]{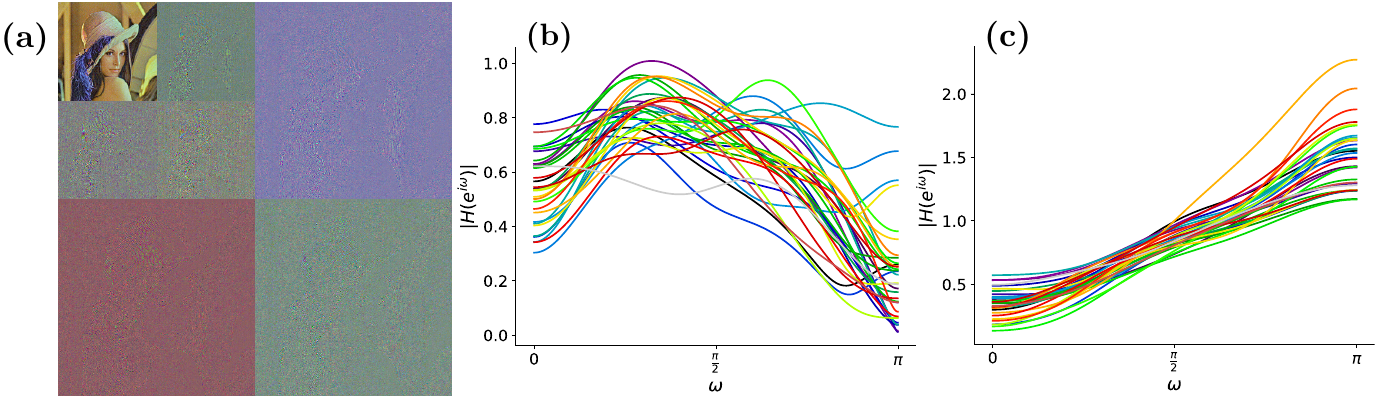}}
\caption{(a). Results from two iterations of the learned LWT on the Lena image ($512\times512$). Plot (b) and (c) are frequency responses of learned LP filters and HP filters, respectively.}
\label{fig:fir}
\end{center}
\end{figure*}

\subsection{Prior distribution and optimization}\label{sec:train}
Using the 3rd connection mentioned in Sec.~\ref{sec:1d}, we build typical distributions for high/low-frequency coefficients as priors to optimize the LWT flow.

As in Sec.~\ref{sec:1d}, the factored-out variables of all iterations should be from uncorrelated independent distributions for them to be high-frequency coefficients.
For the variables remaining after the last iteration, as no further transformations are there to add correlations to them,
they should be from a correlated distribution for them to be low-frequency coefficients. Here, we use a mixture of distributions~\cite{NIPS2016_ddeebdee, salimans2017pixelcnn++} for these last variables.

The overall prior distribution is as follows,
\begin{equation}
  \prod_{j,n} \text{P}(\textbf{H}_{n}^j | \mu_j, \sigma_j) \cdot \sum_i \pi_i \text{P}(\textbf{L}_N|\boldsymbol{M}_i, \boldsymbol{\Sigma}_i)
  \label{eq:prior1}
\end{equation}
where $\textbf{H}_{n}^j$ are high-frequency parts from $n$-th iteration. For 2D LWT, $\textbf{H}_n^j = \{\textbf{H}_n^a,\textbf{H}_n^b,\textbf{H}_n^c\}$; and for 1D LWT, there is only one high-frequency part $\textbf{H}^a_n$. The $\mu_j$, $\sigma_j$ are learnable scalar parameters of the independent distributions.
For the mixture distribution, the $\boldsymbol{M}_i$ and $\boldsymbol{\Sigma}_i$ are learnable high-dimensional parameters of the $i$-th distribution, and $\pi_i$ is the weight of the $i$-th distribution.


For performance, one can make the parameters of high-frequency distributions depend on the low-frequency part of the same iteration. This changes Eq.~(\ref{eq:prior1}) into
\begin{equation}
  \prod_{j, n} \text{P}(\textbf{H}_{n}^j | \boldsymbol{\mu}_j(\textbf{L}_n), \boldsymbol{\sigma}_j(\textbf{L}_n)) \cdot \sum_i \pi_i \text{P}(\textbf{L}_N|\boldsymbol{M}_i, \boldsymbol{\Sigma}_i)
  \label{eq:prior2}
\end{equation}
where $\boldsymbol{\mu}_j(\cdot)$ and $\boldsymbol{\sigma}_j(\cdot)$ are learnable functions that evaluate parameters for the high-frequency distributions.
And $\textbf{L}_n$ is the $n$-th iteration's low-frequency part. Note that, this doesn't violate the uncorrelation condition of high-frequency variables, \ie, the mutual information can still be zero (Appx.~\ref{app:mutualinfo}).

For optimization, one minimizes the negative log-likelihood (NLL) of the transformed variables ($\textbf{H}^j_n$ and $\textbf{L}_N$) on the prior distribution. This gives these transformed variables high/low-frequency features and makes the coupling layer an LP/HP filter pair.
Also,
as LWT has Jacobian determinant of $1$,
one can see from Eq.~(\ref{eq:flowProbability}) that this loss function simultaneously maximizes the sample probability.
Thus, by only optimizing NLL, we unify these two optimization goals.
One can also use the NLL term as a regulation in the overall task loss function for using LWT in certain tasks.

\section{Architecture details}\label{sec:arch}

For the following experiments, we define an LWT for 2D data.
The overall architecture is like Fig.~\ref{fig:overall}(b), and we perform the iterations until the low-frequency variables $\textbf{L}_N$ is a $2\times2$ matrix,
\ie, the $N = \log_2(d) - 2$, where $d$ is the side length of the 2D data and $d\ge 4$.
The coupling layer and factor-out scheme we use is the one in Fig.~\ref{fig:partition}(b), and we use $K$ layers of coupling.
 For priors, we set the parameters of the high-frequency distributions to depend on the low-frequency variables (Eq.~(\ref{eq:prior2})).

\textbf{Sub-networks:} For $t_k(\cdot)$ in Eq.~(\ref{eq:2dNICE}), we use convolutional neural networks (CNN) with input channels three times the output channels. We concatenate the three matrices $\{\textbf{H}_n^a, \textbf{H}_n^b,\textbf{H}_n^c\}$ along the channel dimension, input it into the CNN, and transform it into an output with the same size of $\textbf{L}_n$.
For $\boldsymbol{\mu}_j(\cdot)$, we use one CNN instead of three $\boldsymbol{\mu}_j(\cdot)$, \ie, we use a CNN with output channels three times the input channels, and separate along the output's channel dimension.
We also set up $\boldsymbol{\sigma}_j(\cdot)$ the same way.

\textbf{Discrete variables:} As the following experiments are on image data, we define the LWT on discrete variables, and the input variables $\in [0, 255]$.
For these discrete data to work with $t_k(\cdot)$, $\boldsymbol{\mu}(\cdot)$ and $\boldsymbol{\sigma}(\cdot)$ sub-networks, the inputs are subtracted with $128$ and divided by $255$ to convert them into $[-1/2, 1/2)$. Then, for outputs of these networks, we invert them back to $[0, 255]$, and rounded to the nearest integer. This rounding is approximately differentiable~\cite{hoogeboom19_integ_discr_flows_lossl_compr}.

\textbf{Discrete priors:} The distributions of high-frequency variables of all iterations are parameterized discrete logistic distributions. The distribution of the last iteration's low-frequency part is a mixture of discrete logistic distributions with learnable parameters and weights~\cite{salimans2017pixelcnn++}.

\begin{table*}[ht!]
  \caption{Lossless compression scores (BPD) on CIFAR-10, ImageNet32/64, and ImageNet. The scores in parenthesis are theoretical.
  }
\begin{center}
      \label{tab:compression}
      \begin{tabular}{lcccc}
        \toprule
        Algorithm/Model                                           & CIFAR-10    & ImageNet32    & ImageNet64  & ImageNet\\
        \midrule
        PNG~\cite{boutell1997png}                                 &5.87         &6.39         &5.71        & 4.43\\
        JPEG-2000~\cite{rabbani2002jpeg2000}                      &5.20         &6.48         &5.10        & 4.42\\
        \midrule
        LBB~\cite{ho19_compr_with_flows_via_local}                &3.12 (3.12)  &3.88 (3.87)  &3.70 (3.70) & --  \\
        IDF~\cite{hoogeboom19_integ_discr_flows_lossl_compr}      &3.34 (3.32)  &4.18 (4.15)  &3.90 (3.90) & --   \\
        IDF++~\cite{berg20_idf}                                   &3.26 (3.24)  &4.12 (4.10)  &3.81 (3.81) & -- \\
        \midrule
        LWT (this work)                                            &3.11 (3.10)  &3.50 (3.49)  &3.21 (3.20) &2.99 (2.95)   \\
        \bottomrule
      \end{tabular}
\end{center}
\end{table*}

\section{Numerical analysis}\label{sec:analysis}

In this section, a numerical analysis of the learned transformation is given. The LWT is trained on ImageNet64~\cite{chrabaszcz2017downsampled} via minimizing the NLL. For detailed hyper-parameters, see Appx.~\ref{app:hyper}.

\textbf{Two-iteration plot}: In Fig.~\ref{fig:fir}(a), we perform two iterations of the learned LWT on the Lena image. Then, following Fig.~\ref{fig:overall}(c), we concatenate high-frequency parts from the $0$-th iteration $\{\textbf{H}_{0}^a, \textbf{H}_{0}^b, \textbf{H}_{0}^c\}$, high-frequency parts from the $1$-th iteration $\{\textbf{H}_{1}^a, \textbf{H}_{1}^b, \textbf{H}_{1}^c\}$, and the low-frequency part from the $1$-th iteration $\textbf{L}_1$ into a 2D image.
One can see that the low-frequency part contains non-local information and can be viewed as a blurred signal, while high-frequency parts only contain local details, such as edges and noises. While the optimization goal expels the long-range correlations from high-frequency parts, we want to emphasize that the existence of local details is guaranteed by the invertibility of flows (Appx.~\ref{app:edgeplot}).
Note that, the LWT is learned on ImageNet64 and successfully generalized to Lena image, the two even don't share the same dimensions. For visualization, normalization is performed, see Appx.~\ref{app:normal}.
\textbf{Frequency response}: For a traditional linear filter, a way of viewing its behavior is to plot its frequency response~\cite{oppenheim1999discrete}. For the learned non-linear transformation, we view the gradient matrix of one iteration as the transformation matrix. According to Eq.~(\ref{eq:1Dwavelet}), one can get each element forming the matrix $\textbf{W}$ from the gradients. Then we divide this transformation matrix into LP/HP filter rows according to variable partition in Fig.~\ref{fig:partition}(b). As we use a random sample from ImageNet64, the image has $64$ rows, so there are $32$ LP filters and $32$ HP filters. In Fig.~\ref{fig:fir}(b) and (c), we plot the frequency responses of these learned LP and HP filters, respectively. Details of performing this plot are in Appx.~\ref{app:response}.
One can see from the frequency responses, the learned LWT demonstrates classic LP/HP filter behaviors.

\section{Lossless compression}\label{sec:experiment}

In this section, one application of the LWT is demonstrated: the lossless compression of images. We follow the mature designs of JPEG-2000~\cite{rabbani2002jpeg2000} which is a wavelet-based algorithm. One benefit of compatibility with JPEG-2000 is that our proposed method is easily adaptable for industrial applications.
More details about hyper-parameters and model setups can be found in Appx.~\ref{app:hyper}.

\subsection{Lossless compression scores}\label{sec:compression}

Lossless compression is also a natural application for normalizing flows~\cite{hoogeboom19_integ_discr_flows_lossl_compr, berg20_idf}: the optimization goal of minimizing NLL happens to be minimizing the Shannon’s bound of lossless compression size (Shannon entropy)~\cite{shannon1948mathematical}:
\begin{equation}
  \mathbb{E}_{\mathbf{x}} \rm Len\left(Code(x)\right) \geq H(P_{\mathcal{X}}) = -\mathbb{E}_{\mathbf{x}} \log P_{\mathcal{X}}(x)
  \label{eq:shannon}
\end{equation}
So we train an LWT flow by minimizing the NLL, which also optimizes LWT for the compression task.

To use learned LWT in lossless compression. One transforms the images using this LWT, as in Fig.~\ref{fig:overall}(b), and compresses all the resulting variables by entropy encoding algorithms~\cite{huffman1952method, duda09_asymm_numer_system} according to each variable's probability. These probabilities can be estimated using the prior distribution. Here we use ANS~\cite{duda09_asymm_numer_system} as the entropy encoding algorithm. For details about ANS, see Appx.~\ref{app:rans}.


In Table~\ref{tab:compression}, we provide compression score comparison of different algorithms and models on CIFAR-10~\cite{krizhevsky2009learning}, ImageNet32/64~\cite{chrabaszcz2017downsampled} and ImageNet~\cite{ILSVRC15} datasets. The compression scores are the bits per dimension (BPD), and are evaluated on the test sets.
For uncompressed image data, the BPD should be $8$.
One can see the LWT achieves SOTA scores.
Moreover, as LWT shares parameters, it's smaller than the previous SOTA model (the LBB), its parameter saving file is $10 \times$ smaller on ImageNet64 training and $20\times$ smaller on CIFAR-10. The share of parameters also relaxes the constraint of samples be of the same size. One can perform more iterations for bigger sizes. In this way, one can train or evaluate LWT on datasets with varying sample sizes, \eg, the original ImageNet. 


\subsection{Generalization test}\label{sec:generalization}
\begin{table}[!bp]
  \caption{Achieved BPD scores on different datasets from trained on.}
\begin{center}
      \label{tab:migration}
      \begin{tabular}{lcccr}
        \toprule
        &  \multicolumn{2}{c}{Train on} \\
        \cmidrule(lr){2-3}
        Evaluate on                                              & ImageNet32    & ImageNet64  \\
        \midrule
        ImageNet32                                               & 3.50          & 3.50        \\
        ImageNet64                                               & 3.24          & 3.21        \\
        ImageNet                                                 & 3.03            & 2.99        \\
        \bottomrule
      \end{tabular}
\end{center}
\end{table}
For this part, we test the model's generalization ability.
Due to the parameter-sharing scheme, the learned transformation poses no constraint on the input size. So, one can naturally evaluate the ImageNet64-trained model on the ImageNet32 dataset, and so on. Moreover, parameter sharing also allows one to evaluate on the original ImageNet which has images of different sizes.
As shown in Table~\ref{tab:migration}, the LWT has substantial generalization ability.
One can see that the LWT flows evaluate approximately the same score independent of what resolutions they are initially trained on.
This implies one can train LWT on low-resolution datasets, and use it on high-resolution images. 

\subsection{Progressive loading}\label{sec:loading}

Another useful feature of JPEG-2000 is the progressive loading.
This is one of the improvements of JPEG-2000 over its predecessor~\cite{charrier1999jpeg2000, rabbani2002jpeg2000}.
When transporting images, we first transport the low-frequency parts, the algorithm can fill out the high-frequency parts to give previews.

\begin{figure}[!tbp]
\begin{center}
\centerline{\includegraphics[width=0.9\columnwidth]{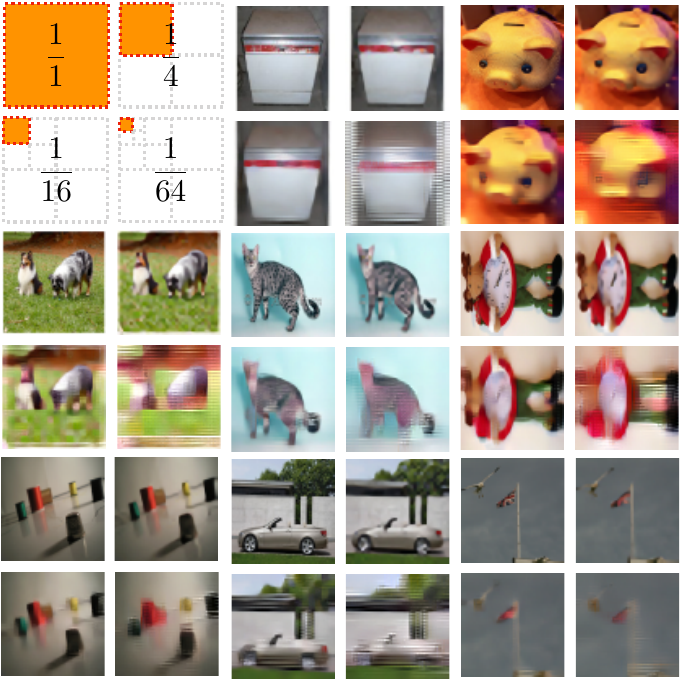}}
\caption{Progressive loading from $1/64$, $1/16$, and $1/4$ the original size back to the original size. The parts marked with orange are original (the low-frequency variables given). The images are from ImageNet64, and the used LWT is trained on it.}
\label{fig:proloading1}
\end{center}
\end{figure}

To do progressive loading using the learned LWT, a low-frequency part of a certain $n$-th iteration is given, \ie, $\textbf{L}_n$.
As we set high-frequency parts $\{\textbf{H}_{n}^a, \textbf{H}_{n}^b,\textbf{H}_{n}^c\}$ to be dependent on $\textbf{L}_n$, one can sample $[\widetilde{\textbf{H}}_{n}^a, \widetilde{\textbf{H}}_{n}^b, \widetilde{\textbf{H}}_{n}^c]$ by drawing from discrete logistic distribution $\text{P}(\textbf{H}_{n}^j | \boldsymbol{\mu}_j(\textbf{L}_n), \boldsymbol{\sigma}_j(\textbf{L}_n))$.
Then, an inverse LWT is performed, to convert these variables to an approximate low-frequency part $\widetilde{\textbf{L}}_{n-1}$ of $(n-1)$-th iteration. This is repeated until the desired size is reached. A demonstration of progressive loading from $1/64$, $1/16$, and $1/4$ the original size back to the original size is in Fig.~\ref{fig:proloading1}.
Additionally, one can perform upsampling/super-resolution in a similar way, more details in Appx.~\ref{app:sr}.


\section{Related work}\label{sec:related}
\textbf{Normalizing flow model}: This model can be viewed as a novel type of normalizing flow model. Comparing with the commonly used realNVP model~\cite{dinh16_densit_estim_using_real_nvp}, this model updates variables using a different coupling, the generalized lifting coupling, as in Fig.~\ref{fig:partition}. This model also has a factor-out layer, but follows the wavelet transformation's downsampling.
More importantly, LWT shares parameters along different iterations/levels, repeating the same transformation/filtering. This share of parameter makes LWT much smaller than usual flow models, and lift the constraint of inputs having the same dimension.

\noindent\textbf{Learnable filter kernel}: There have been attempts to propose learnable filter kernels for wavelets~\cite{khan2018learning, balestriero2018spline, ravanelli2018interpretable, cosentino2020learnable, gilles2013empirical}.
The difference is that we propose inherent connections between flows and wavelets, which allow us to naturally generalize the entire wavelet framework to deep learning models (Fig.~\ref{fig:overall}) not just using neural networks as filters. These findings give wavelet filtering a probability/information background, allowing one to optimize filters using probability distributions that fits deep learning approaches naturally. Also, our model allows end-to-end training with the task goal built into the loss function, which usually gives better optimization results (\eg, SOTA compression results shown in Sec.~\ref{sec:compression}).

\noindent\textbf{Deep-learning lossless compression}: We demonstrate the application of image lossless compression. There have been works that connect flow models with lossless compression~\cite{hoogeboom19_integ_discr_flows_lossl_compr, berg20_idf}. The difference between them and Sec.~\ref{sec:experiment} is the use of LWT, which allows us to introduce mature designs from JPEG-2000. Also, the wavelet nature of LWT makes the presented method naturally generalizes to different input sizes, which is crucial for a deep-learning compression method to be widely applicable.

\section{Summary}\label{sec:conclusion}

In this work, a novel kind of normalizing flow is proposed. This model extends linear wavelet transformation and constitutes a scheme of learnable non-linear wavelet transformation.
To the best of our knowledge, this is the first work that proposes this inherent connection between flows and wavelet transformations.

As demonstrated in the paper, the proposed model can learn a wavelet transformation. Compare to traditional linear wavelets, this model is non-linear and optimized for certain tasks and data, thus has better expressive power. As a novel normalizing flow, its wavelet-like behavior grants it efficiency and flexibility: the share of parameters makes the model small in size and efficient in computation; parameter sharing also lifts the dimensional constraint and makes the model generalize well.

We also demonstrate the model's power in image compression, it achieves SOTA scores and substantial generalization ability. Additionally, as a wavelet, it's compatible with JPEG-2000, which means one can use mature designs from JPEG-2000 and easily adapt it for industrial applications.
Other potential usage of this model include super-resolution, ICA tasks, and multiple domain-specific cases that traditionally use wavelets.

\nocite{SM, Code}

\bibliographystyle{icml2022}
\bibliography{ref}

\appendix

\clearpage
\pagebreak
\onecolumn

\renewcommand{\theequation}{\thesection.\arabic{equation}}
\renewcommand{\thetable}{\thesection.\arabic{table}}
\renewcommand{\thefigure}{\thesection.\arabic{figure}}

\setcounter{table}{0}
\setcounter{equation}{0}
\setcounter{figure}{0}
\section{Theoretical details}\label{app:glminit}

\subsection{Generalized lifting scheme}\label{app:glm}

The $8\times 8$ transformation matrix $\mathbf{W}$ for the Haar wavelet~\cite{harr1910theorie} can be formulated as follows,
\begin{equation}
    \label{eq:harr}
  \mathbf{W}_{\rm{haar}} =
  \left[
    \begin{matrix}
       1/2 & 1/2 & 0 & 0 & 0 & 0 & 0 & 0 \\
        0 & 0 & 1/2 & 1/2 & 0 & 0 & 0 & 0 \\
        0 & 0 & 0 & 0 & 1/2 & 1/2 & 0 & 0 \\
        0 & 0 & 0 & 0 & 0 & 0 & 1/2 & 1/2 \\
        -1 & 1 & 0 & 0 & 0 & 0 & 0 & 0 \\
        0 & 0 & -1 & 1 & 0 & 0 & 0 & 0 \\
        0 & 0 & 0 & 0 & -1 & 1 & 0 & 0 \\
        0 & 0 & 0 & 0 & 0 & 0 & -1 & 1 
    \end{matrix}
  \right]
\end{equation}
The corresponding transformation as follows, as in Eq.~(\ref{eq:1Dwavelet})

\begin{equation}
  \label{eq:harr2}
  \left[ \begin{matrix} 1/2 & 1/2 & 0 & 0 & 0 & 0 & 0 & 0 \\ 0 & 0 & 1/2 & 1/2 & 0 & 0 & 0 & 0 \\ 0 & 0 & 0 & 0 & 1/2 & 1/2 & 0 & 0 \\ 0 & 0 & 0 & 0 & 0 & 0 & 1/2 & 1/2 \\ -1 & 1 & 0 & 0 & 0 & 0 & 0 & 0 \\ 0 & 0 & -1 & 1 & 0 & 0 & 0 & 0 \\ 0 & 0 & 0 & 0 & -1 & 1 & 0 & 0 \\ 0 & 0 & 0 & 0 & 0 & 0 & -1 & 1  \end{matrix}\right] \cdot \left[\begin{matrix} x_0 \\ x_1 \\ x_2 \\ x_3 \\ x_4 \\ x_5 \\ x_6 \\ x_7 \end{matrix}\right]
  = \left[ \begin{matrix}  x_0/2 + x_1/2 \\ x_2/2 + x_3/2 \\x_4/2 + x_5/2 \\ x_6/2 + x_7/2\\ -x_0 + x_1 \\ -x_2 + x_3 \\ -x_4 + x_5 \\ -x_6 + x_7 \end{matrix}\right] = \textbf{z}
\end{equation}

The low-pass filter of the Haar wavelet is defined as $(1/2, 1/2)$, and the high-pass filter is $(-1,1)$.
For the resulting vector of Eq.~(\ref{eq:harr2}), the upper half is the low-pass result, and the lower half is the high-pass result.

Another way of performing Haar wavelet is through lifting~\cite{sweldens1998lifting}.
We first separate input vector $\textbf{x}$ into two parts: the odd-index part $\textbf{x}_o$ and the even-index part $\textbf{x}_e$. \ie,
\begin{equation}
  \label{eq:parts}
    \begin{split}
    \textbf{x}_o &= [x_1, x_3, x_5, x_7, \cdots]\\
    \textbf{x}_e &= [x_0, x_2, x_4, x_6, \cdots] 
  \end{split}
\end{equation}
Then the lifting for Haar can be written as
\begin{equation}
  \label{eq:originalharrlifting}
  \begin{split}
    \textbf{z}_o &= \textbf{x}_o - \textbf{x}_e\\
    \textbf{z}_e &= \textbf{x}_e + \lfloor \textbf{z}_o/2 \rceil 
  \end{split}
\end{equation}
And one can write this lifting of Haar wavelet into two layers of NICE coupling.
The first layer of coupling as follows, and we save the results into two intermediate vectors:
\begin{equation}
  \label{eq:highVar}
  \begin{split}
    \textbf{y}_o &= \textbf{x}_o -  \textbf{x}_e \\
    \textbf{y}_e &= \textbf{x}_e
  \end{split}
\end{equation}
And then, the final result can be derived from another layer of coupling
\begin{equation}
  \label{eq:lowVar}
  \begin{split}
    \textbf{z}_o &= \textbf{y}_o \\
    \textbf{z}_e &= \textbf{y}_e + \lfloor \textbf{y}_o/2 \rceil 
   \end{split}
 \end{equation}

 The high-pass result is the $\textbf{z}_o$ and the low-pass result is the $\textbf{z}_e$.
 
Eq.~(\ref{eq:highVar}) and Eq.~(\ref{eq:lowVar}) can be formulated as two typical NICE~\cite{dinh2015nice} coupling layers,
as long we can express the change terms (\ie, the $-\textbf{x}_e$ and $\lfloor \textbf{y}_o/2 \rceil$ terms) using the $t$ transformation of Eq.~(\ref{eq:vanillaNICE}). In this way, we can view the NICE coupling layer with even-odd separation as a generalized lifting scheme.

As an additional example, the LeGall wavelet~\cite{le1988sub} has the following transformation matrix:
\begin{equation}
  \label{eq:legall}
  \mathbf{W}_{\rm{legall}} =
  \left[
    \begin{matrix}
      3/4 & 1/2 & -1/4 & 0 & 0 & 0 & 0 & 0 \\
        -1/8 & 1/4 & 3/4 & 1/4 & -1/8 & 0 & 0 & 0 \\
        0 & 0 & -1/8 & 1/4 & 3/4 & 1/4 & -1/8 & 0 \\
        0 & 0 & 0 & 0 & -1/8 & 1/4 & 5/8 & 1/4 \\
        -1/2 & 1 & -1/2 & 0 & 0 & 0 & 0 & 0 \\
        0 & -1/2 & 1 & -1/2 & 0 & 0 & 0 & 0 \\
        0 & 0 & 0 & -1/2 & 1 & -1/2 & 0 & 0 \\
        0 & 0 & 0 & 0 & 0 & 0 & -1 & 1 
    \end{matrix}
  \right]
\end{equation}

And the corresponding transformation, as in Eq.~(\ref{eq:1Dwavelet})

\begin{multline}
  \label{eq:legall2}
  \left[ \begin{matrix} 3/4 & 1/2 & -1/4 & 0 & 0 & 0 & 0 & 0 \\ -1/8 & 1/4 & 3/4 & 1/4 & -1/8 & 0 & 0 & 0 \\ 0 & 0 & -1/8 & 1/4 & 3/4 & 1/4 & -1/8 & 0 \\ 0 & 0 & 0 & 0 & -1/8 & 1/4 & 5/8 & 1/4 \\ -1/2 & 1 & -1/2 & 0 & 0 & 0 & 0 & 0 \\ 0 & -1/2 & 1 & -1/2 & 0 & 0 & 0 & 0 \\ 0 & 0 & 0 & -1/2 & 1 & -1/2 & 0 & 0 \\ 0 & 0 & 0 & 0 & 0 & 0 & -1 & 1  \end{matrix}\right] \cdot \left[\begin{matrix} x_0 \\ x_1 \\ x_2 \\ x_3 \\ x_4 \\ x_5 \\ x_6 \\ x_7 \end{matrix}\right]\\
  = \left[ \begin{matrix} -x_2/8 + x_1/4 + 3x_0/4 + x_1/4 - x_2/8 \\ -x_0/8 + x_1/4 + 3x_2/4 + x_3/4 - x_4/8 \\ -x_2/8 + x_3/4 + 3x_4/4 + x_5/4 - x_6/8 \\ -x_4/8 + x_5/4 + 3x_6/4 + x_7/4 - x_6/8 \\ -x_0/2 + x_1 - x_2/2 \\ -x_2/2 + x_3 - x_4/2 \\ -x_4/2 + x_5 - x_6/2 \\ -x_6/2 + x_7 - x_6/2 \end{matrix}\right] = \textbf{z}
\end{multline}

The low-pass filter is defined as $(-1/8, 1/4, 3/4, 1/4, -1/8)$, and the high-pass filter is $(-1/2,1,-1/2)$.

And it can also be formulated using the generalized lifting scheme, or NICE coupling layers. We first separate the odd-index variables from the even-index one, \ie,
\begin{equation}
  \label{eq:legallpart}
      \begin{split}
    \textbf{x}_o &= [x_1, x_3, x_5, x_7, \cdots]\\
    \textbf{x}_e &= [x_0, x_2, x_4, x_6, \cdots] 
  \end{split}
\end{equation}
Then, perform
\begin{equation}
  \label{eq:legalllifting}
  \begin{split}
    y_{o,i} &= x_{o,i} - \lfloor (x_{e,i} + x_{e,i+1})/2 \rceil \\
    y_{e,i} &= x_{e,i}
  \end{split}
  ~~
  \text{ and }
  ~~
  \begin{split}
    z_{o,i} &= y_{o,i} \\
    z_{e,i} &= y_{e,i} + \lfloor (y_{o,i} + y_{o,i-1})/4 \rceil 
   \end{split}
 \end{equation}
 as two layers of NICE coupling. Subscript $i$ here stands for indexes of variables in the vectors. The resulting $\textbf{z}_o$ is the high-pass result, and the $\textbf{z}_e$ is the low-pass result.

\subsection{Initialize 1D LWT to be LeGall}\label{app:init}
Eq.~(\ref{eq:legalllifting}), Eq.~(\ref{eq:highVar}), and Eq.~(\ref{eq:lowVar}) can be used to initialize LWTs. One has to convert the change terms (\ie, the $-\lfloor (x_{e,i} + x_{e,i+1})/2 \rceil$, $\lfloor (y_{o,i} + y_{o,i-1})/4 \rceil$, $-\textbf{x}_e $, and $\lfloor \textbf{y}_o / 2 \rceil$ terms) into convolutional neural networks (CNNs), and use these CNNs to initialize the $t$ transformations of Eq.~(\ref{eq:1Dcoupling}).

One obstacle of converting the change terms is that the activation function ReLu of CNN eliminates negative values. To circumvent this, one can use hidden-layer CNNs with more than $6$ channels, and distribute opposite values of the input on the other $3$ channels. Then at the last layer, one removes the last ReLu activation, and converts these channels back into a $3$-channel output.

For example, to initialize a LeGall wavelet LWT, we write the coupling, Eq.~(\ref{eq:1Dcoupling}), as follows.
\begin{equation}
  \label{eq:NICEinit}
  \begin{split}
    [\textbf{L}_n, \textbf{H}_{n}^a]_{k+1} &= [\textbf{L}_n, \textbf{H}_{n}^a - \code{layer1}(\textbf{L}_{n})]_{k} \\
    [\textbf{L}_n, \textbf{H}_{n}^a]_{k+2} &= [\textbf{L}_n + \code{layer2}(\textbf{H}_{n}^a), \textbf{H}_{n}^a]_{k+1} 
  \end{split}
 \end{equation}
 \code{layer1} and \code{layer2} are the $t$ transformations of Eq.~(\ref{eq:1Dcoupling}), and can be initialized to be change terms in Eq~(\ref{eq:legalllifting}).
 They are 1D convolutional networks with channel change $3\rightarrow 10\rightarrow 10 \rightarrow 3$.
 And \code{layer1} has a structure of \\
 \renewcommand{\labelenumi}{(\arabic{enumi})}
 \begin{enumerate}
 \item 1D replication padding two values on the right hand side.
 \item Conv1d(input\_channel=3, output\_channel=10, kernel\_size=3, stride=1)
 \item ReLU
 \item Conv1d(input\_channel=10, output\_channel=10, kernel\_size=3, stride=1, padding=1,\\ padding\_mode=replicate)
 \item ReLU
 \item Conv1d(input\_channel=10, output\_channel=3, kernel\_size=3, stride=1, padding=1,\\ padding\_mode=replicate)
 \end{enumerate}
 The first $(6,3,3)$ part of the weight for the first Conv1d is
 \begin{multline}
   \left[
     \left[\begin{matrix}
       0.5&  0.5&  0\\
       0&  0&  0\\
       0&  0&  0
     \end{matrix}\right]~,
     ~~
     \left[\begin{matrix}
       0&  0&  0\\
       0.5&  0.5&  0\\
       0&  0&  0       
     \end{matrix}\right]~,
     ~~
     \left[\begin{matrix}
       0&  0&  0\\
       0&  0&  0\\   
       0.5&  0.5&  0
     \end{matrix}\right]~,\right.
     ~~\\
     \left.\left[\begin{matrix}
         -0.5&  -0.5&  0\\
         0&  0&  0\\
         0&  0&  0
       \end{matrix}\right]~,
     ~~
     \left[\begin{matrix}
       0&  0&  0\\
       -0.5&  -0.5&  0\\
       0&  0&  0       
     \end{matrix}\right]~,
     ~~
     \left[\begin{matrix}
       0&  0&  0\\
       0&  0&  0\\   
       -0.5&  -0.5&  0
     \end{matrix}\right]     
   \right]
 \end{multline}
The first $(6, 6, 3)$ part of the weight for the second Conv1d is
\begin{equation}
   \left[
     \left[\begin{matrix}
         0& 1& 0\\
         0& 0& 0\\
         0& 0& 0\\
         0& 0& 0\\
         0& 0& 0\\
         0& 0& 0
     \end{matrix}\right]~,
     ~~
     \left[\begin{matrix}
         0& 0& 0\\
         0& 1& 0\\
         0& 0& 0\\
         0& 0& 0\\
         0& 0& 0\\
         0& 0& 0
     \end{matrix}\right]~,
     ~~
     \left[\begin{matrix}
         0& 0& 0\\
         0& 0& 0\\
         0& 1& 0\\
         0& 0& 0\\
         0& 0& 0\\
         0& 0& 0
       \end{matrix}\right]~,
     ~~
     \left[\begin{matrix}
         0& 0& 0\\
         0& 0& 0\\
         0& 0& 0\\
         0& 1& 0\\
         0& 0& 0\\
         0& 0& 0
     \end{matrix}\right]~,
     ~~
     \left[\begin{matrix}
         0& 0& 0\\
         0& 0& 0\\
         0& 0& 0\\
         0& 0& 0\\
         0& 1& 0\\
         0& 0& 0
     \end{matrix}\right]~,
     ~~
     \left[\begin{matrix}
         0& 0& 0\\
         0& 0& 0\\
         0& 0& 0\\
         0& 0& 0\\
         0& 0& 0\\
         0& 1& 0
     \end{matrix}\right]     
   \right]
\end{equation}
The first $(3, 6, 3)$ part of the weight for the third Conv1d is
\begin{equation}
   \left[
     \left[\begin{matrix}
         0&  1&  0\\
         0&  0&  0\\
         0&  0&  0\\
         0& -1& 0\\
         0& 0& 0\\
         0& 0& 0
     \end{matrix}\right]~,
     ~~
     \left[\begin{matrix}
         0&  0&  0\\
         0&  1&  0\\
         0&  0&  0\\
         0& 0& 0\\
         0& -1& 0\\
         0& 0& 0
     \end{matrix}\right]~,
     ~~
     \left[\begin{matrix}
         0&  0&  0\\
         0&  0&  0\\
         0&  1&  0\\
         0& 0& 0\\
         0& 0& 0\\
         0& -1& 0
     \end{matrix}\right]
   \right]
\end{equation}

And the rest of the weights and all bias are set to zeros.

\code{layer2} is defined similarly, with the only difference being changing the first 1D replication padding to the left-hand side. And the weights and bias are initialized the same as \code{layer1}, excepting the first $(6,3,3)$ part of the weight for the first Conv1d is initialized as
 \begin{multline}
   \left[\left[\begin{matrix}
         0&  0.25&  0.25\\
         0&  0&  0\\
         0&  0&  0
       \end{matrix}\right]~,
     ~~
     \left[\begin{matrix}
         0&  0&  0\\
         0&  0.25&  0.25\\
         0&  0&  0       
     \end{matrix}\right]~,
     ~~
     \left[\begin{matrix}
       0&  0&  0\\
       0&  0&  0\\
       0&  0.25&  0.25
     \end{matrix}\right]~,\right.
     ~~ \\
     \left.\left[\begin{matrix}
         0&  -0.25&  -0.25\\
         0&  0&  0\\
         0&  0&  0
       \end{matrix}\right]~,
     ~~
     \left[\begin{matrix}
         0&  0&  0\\
         0&  -0.25&  -0.25\\
       0&  0&  0       
     \end{matrix}\right]~,
     ~~
     \left[\begin{matrix}
       0&  0&  0\\
       0&  0&  0\\
       0&  -0.25&  -0.25
     \end{matrix}\right]
                  \right]
\end{multline}

\subsection{Uncorrelation condition}\label{app:mutualinfo}

The mutual information between random Gaussian variables only concerns the off-diagonal elements of the correlation matrix.
\eg, for two correlated Gaussian variables $X_1$ and $X_2$ with a correlation matrix like
\begin{equation}
  \label{eq:corrmatrix}
  \boldsymbol{\Sigma} =
  \left[
  \begin{matrix}
    \sigma_1^2 & \rho\sigma_1\sigma_2 \\
    \rho\sigma_1\sigma_2 & \sigma_2^2 
  \end{matrix}
  \right]
\end{equation}
The mutual information between the two variables is~\cite{gel1959calculation}
\begin{equation}
  \label{eq:mutualinformation}
  \text{I}(X_1; X_2) = -\frac{1}{2}\log(1-\rho^2)
\end{equation}
When the off-diagonal elements are zero, \ie, $\rho = 0$, one can see that the mutual information is zero.

So, to protect the uncorrelation condition, one can set the off-diagonal elements of the correlation matrix to be zero when setting up the prior distributions.

\setcounter{table}{0}
\setcounter{equation}{0}
\setcounter{figure}{0}
\section{Experimental details}\label{app:experiment}

\subsection{Training details and hyper-parameters}\label{app:hyper}

The LWT model in use has a \code{repeat} number of coupling layers, \ie, the $K$ in Fig.~\ref{fig:partition}. And the total number of discrete logistic distributions in the mixture is $5$, meaning in Eq.~(\ref{eq:prior2}) the $i\in [0,4]$.

The CNNs for $t_k(\cdot)$, $\boldsymbol{\mu}(\cdot)$ and $\boldsymbol{\sigma}(\cdot)$ networks have a structure like~\cite{kingma18_glow}: the first and last CNN layers have kernels that are $3 \times 3$, and the rest CNNs' kernels are $1\times 1$. We use paddings to make sure the width and length of the input and output are the same.
The number of hidden layers, \ie, the number of $1 \times 1$-kernel CNNs, is a hyper-parameter: the \code{n\_hidden}.
The input channel number for $t_k(\cdot)$ is $9$, for we concatenate three high-frequency parts each with $3$ channels along the channel dimension.
The output channel for $t_k(\cdot)$ is $3$ to fit low-frequency part.
As for the number of channels in the hidden layers, it's a hyper-parameter, \code{hidden\_channel}.
For $\boldsymbol{\mu}(\cdot)$ and $\boldsymbol{\sigma}(\cdot)$, the input channel number is $3$, and the output channel number is $9$ to provide parameters for all three high-frequency parts.
The hidden layers of $\boldsymbol{\mu}(\cdot)$ and $\boldsymbol{\sigma}(\cdot)$ also share the \code{n\_hidden} and \code{hidden\_channel} hyper-parameters.

The optimizer we use is the Adamax~\cite{kingma2014adam}. And the learning rate is computed as in~\cite{hoogeboom19_integ_discr_flows_lossl_compr}, \ie, $lr = lr_{\textrm{base}}\cdot \textrm{decay}^{\textrm{epoch}}$.

In Tab.~\ref{tab:hyperPara}, we give detailed hyper-parameters for each case. All models used in Sec.~\ref{sec:analysis} and Sec.~\ref{sec:experiment} can be found here.

\begin{table*}[!ht]
  \caption{Hyper-parameters for different datasets}
\begin{center}
      \label{tab:hyperPara}
      \begin{tabular}{lccccccc}
        \toprule
        Datasets                               &batchsze    &repeat    &n\_hidden  &hidden\_channel  &lr      &decay       \\
        \midrule
        CIFAR-10                                   &128         &1         &1          &450              &0.001   &0.999        \\
        ImageNet32                                 &64          &3         &3          &450              &0.001   &0.99        \\
        ImageNet64                                 &64          &3         &3          &450              &0.001   &0.99        \\
        ImageNet                                   &256 (varys *) &1 &1 &250 &0.001 &0.99 \\
        \bottomrule
      \end{tabular}
\end{center}
       * The biggest batch size for the original ImageNet is $256$, it varies with sample dimensions to fit into GPU memory.
      
\end{table*}

In Sec.~\ref{sec:experiment}, we follow JPEG-2000, and use \code{uint8} YCbCr datasets. To prepare these \code{uint8} datasets, each sample is multiplied by $255$ and rounded to the nearest \code{uint8} integer.
And we convert RGB color scheme to YCbCr following JPEG-2000 convention~\cite{hamilton2004jpeg}, which formulates as
\begin{equation}
  \label{eq:RGB2YCC}
  \left[
    \begin{matrix}
      Y \\
      Cb \\
      Cr 
    \end{matrix}
  \right]
  =
  \left[
    \begin{matrix}
      0.299& 0.587& 0.114\\
      -0.1687& -0.3313& 0.5\\
      0.5& -0.4187& -0.0813
    \end{matrix}
  \right]
  \cdot
  \left[
    \begin{matrix}
      R\\
      G\\
      B
    \end{matrix}
  \right]
  +
  \left[
    \begin{matrix}
      0\\
      128\\
      128
    \end{matrix}
  \right]
\end{equation}

The dataset CIFAR-10 contains $50000$ images and the test set contains $10000$ images. ImageNet32/64 and ImageNet contain approximately $1250000$ train images and $50000$ test images. The images from the ImageNet dataset have different dimensions, ranging from $8$ to $4096$. For convenience, we reshape the images to the nearest dimensions that are a power of $2$, and we make the images' aspect ratio $1$. The resize function we use is \code{resize()} from PIL~\cite{clark2015pillow} with \code{PIL.Image.BOX} method, following~\cite{chrabaszcz2017downsampled}.

The hardware we use is NVIDIA V100 GPUs, and for the ImageNet dataset, we use one NVIDIA A100. The implementation of this model uses the Pytorch framework~\cite{pytorch19}.

\subsection{Normalization for two iteration plot}\label{app:normal}

For the two iteration plots, we perform normalization to make visualization more clear.
One first subtracts distribution means (the zero points) from each high-frequency part, and then normalizes all parts to be $[0,1]$ \code{float} within each channel to plot.

\subsection{Edges in two iteration plot}\label{app:edgeplot}

One can theoretical proof that edges are presented in the high-frequency part of the two iteration plot (Fig.~\ref{fig:fir}(a)). Flows are continuous bijective mappings. ``Continuous'' means close values are mapped closely (the proposed LWT is on discrete variables, but we use rounding to turn continuous variables into discrete ones, which wouldn’t harm this). ``Bijective'' means different values are mapped differently. If there is an edge (a sudden change) in the input, it must be in the output.

\begin{figure*}[!tbp]
\begin{center}
\centerline{\includegraphics[width=0.4\textwidth]{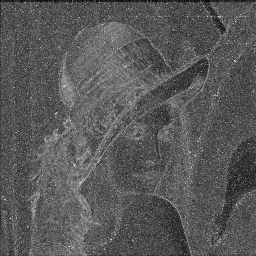}}
\caption{Negative log probability plot of the high-frequency part}
\label{fig:edgeprob}
\end{center}
\end{figure*}

If one requires visually clear edges for the LWT model, we would recommend the use of the probability plot of the high-frequency part where outline points (edges) are assigned unlikely probabilities and trivial points (backgrounds) degenerate to a brand of trivial probabilities. As an example, in Fig.~\ref{fig:edgeprob}, we give the probability plot of one of the high-frequency parts of Fig.~\ref{fig:fir}(a).

\subsection{Frequency response plot}\label{app:response}

As in Eq.~(\ref{eq:legall2}), the gradient matrix contains the low-pass filter $(-1/8, 1/4, 3/4, 1/4, -1/8)$ at rows of the upper half,  and the high-pass filter $(-1/2, 1, -1/2)$ at rows of the lower half. Similarly, the gradient matrix of one iteration of the learned transformation can be used to extract low-pass/high-pass filters. First, one takes the gradient matrix of one output row \wrt the corresponding input row. According to the partition scheme plotted in Fig.~\ref{fig:partition}(b), the even-numbered rows in this gradient matrix are low-pass filters, and the odd-numbered rows are high-pass filters. These row vectors $h$ can then be used to compute the frequency responses $H$ as follows~\cite{oppenheim1999discrete}.
\begin{equation}
  \label{eq:response}
  |H(e^{i\omega})| = \sum_{n=0}^{N}h[n]e^{-i\omega n}
\end{equation}
The $N$ here is the number of values in each row vector. And $\omega$ ranges from $0$ to $\pi$.

\subsection{Entropy encoding and normalizing flow compression model}\label{app:rans}

We give an intuitive explanation of entropy encoding: the encoding process can be viewed as assigning a code string for each character. For a more frequently apparent character, we assign a shorter code. Then the expected length is
\begin{equation}
  \label{eq:expectLen}
  \sum_{\textbf{s}}\text{P}(s)\text{Len}(\text{Code}(s))
\end{equation}
Then, this expectation is lower bounded by Shannon entropy (Eq.~(\ref{eq:shannon})). One straightforward way of achieving this kind of encoding is the Huffman encoding algorithm~\cite{huffman1952method}, which uses a priority queue to sort the probabilities (frequencies) of characters. However, this implementation doesn't have a subtle discern of probability.
One better algorithm is the asymmetric numeral systems algorithm(ANS)~\cite{duda13_asymm_numer_system, duda09_asymm_numer_system}. This algorithm expects a performance close to the Shannon entropy, and similar speed as Huffman encoding. For a state $x$, the character to encode $s$, and corresponding probability $\text{p}(s)\approx l_s/m$, the code $C$ is
\begin{equation}
  \label{eq:ANS}
  C(s, x) = m \lfloor x/l_s \rfloor + b_x + \textrm{mod}(x, l_s)
\end{equation}
where $b_x = \sum_{i=1}^{s-1} l_i$. Then, to decode from state $x$
\begin{equation}
  \label{eq:ANSdecode}
  \begin{split}
  &D(x) = (s, l_s\lfloor x/m\rfloor + \textrm{mod}(x, m) - b_s) \\
  &\text{where $s$ defined as } b_s\ge \textrm{mod}(x, m) < b_{s+1}
  \end{split}
\end{equation}

One key point of entropy encoding is to assess a character's probability. This is where normalizing flows enter the picture, as they mapping the distribution of data into tractable distributions in the latent space. Thus, they can be used to assess the characters' probabilities in the entropy encoding.

Flow establishes a bijective mapping between variables in data space and latent space, \ie, $f(\textbf{s}) = \textbf{z}$. And for each $z$ variable in $\textbf{z}$, one can access its probability using $P_{\mathcal{Z}}(z)$. Then a code can be assigned using Eq.~(\ref{eq:ANS}). This is done for each $z$ to finish the encoding.
As for decoding, we use Eq.~(\ref{eq:ANSdecode}) to decode each $z$ according to $P_{\mathcal{Z}}$. And as flows are invertible, we perform $f^{-1}(\textbf{z})=\textbf{s}$ to get the original data back~\cite{hoogeboom19_integ_discr_flows_lossl_compr}.

\subsection{Upsampling}\label{app:sr}

\begin{figure*}[!tbp]
\begin{center}
\centerline{\includegraphics[width=\textwidth]{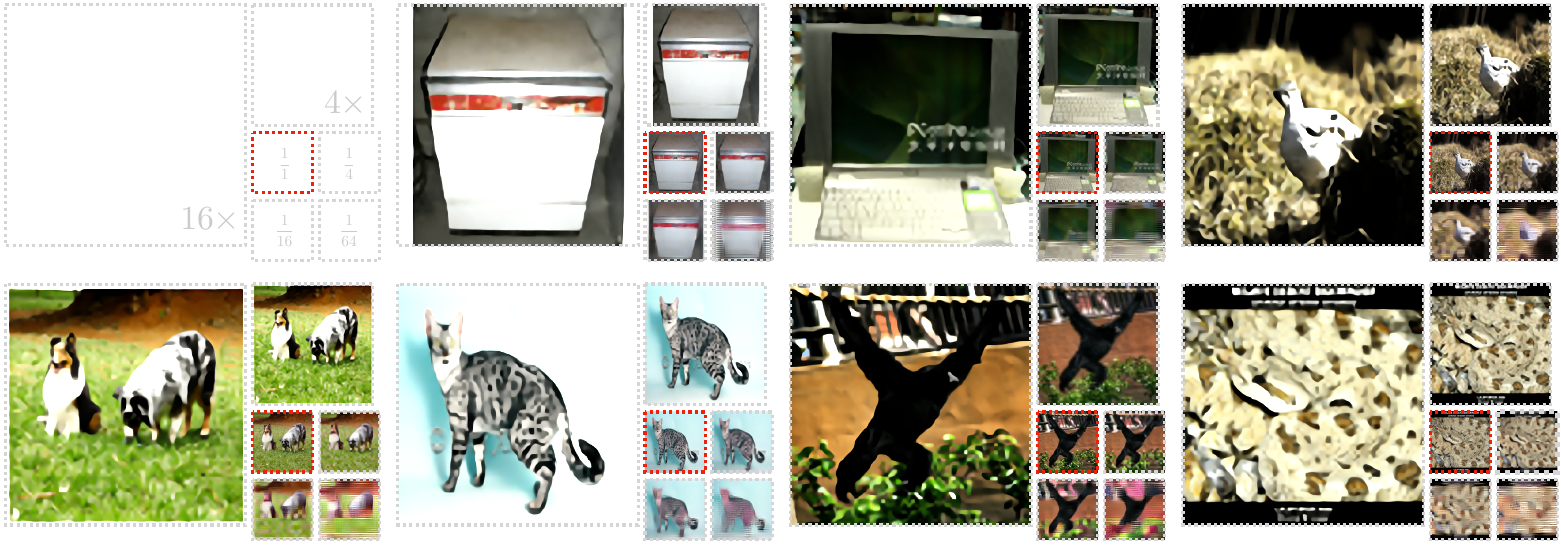}}
\caption{Progressive loading from $1/64$, $1/16$, and $1/4$ original size back to the original size. And super-resolution of original images to $4\times$ and $16\times$ size. The layout is diagramed in the first plot. The images marked with red are the original images. The original images are from ImageNet64, and the model is trained on it.}
\label{fig:proloading}
\end{center}
\end{figure*}

Upsampling/super-resolution is a traditional usage for wavelets, so one can also use LWT. For LWT to perform upsampling/super-resolution, the process is similar to progressive loading in Sec.~\ref{sec:loading}. Only that one is given the original image at the start, and viewed as a low-frequency part of a certain iteration. Then one can sample the corresponding high-frequency parts and perform inverse LWT. This is repeated until the desired size. A demonstration is in Fig.~\ref{fig:proloading}.

One problem that may occur in the upsampling/super-resolution is the color shift. This is assumed to be a training unstable problem. It disappears when models are trained on high-dimension datasets. These high-dimension datasets take more iterations of transformation, and the model has more levels of coupling layers. Thus a better fit for the ideal situation mentioned in Sec.~\ref{sec:train} (low-frequency parts have long-range correlations). While small-dimension datasets often result in learning color shift as a bias.
The solution, other than using a high-dimension dataset, can be using color matching algorithms, such as~\cite{reinhard2001color}. For Fig.~\ref{fig:proloading}, color matching is not used.

\setcounter{table}{0}
\setcounter{equation}{0}
\setcounter{figure}{0}

\section{Pseudocode of 2D LWT}\label{app:pseudocode}

In this section, we present the pseudocode of LWTs using the two 2D coupling schemes mentioned in Sec.~\ref{sec:2d}.

\begin{algorithm}[hb!]
  \SetKwInOut{Input}{input}
  \caption{2D learnable wavelet transformation (scheme 1, with coupling layers diagramed in Fig.~\ref{fig:partition}(b))}
  \label{alg:2dwaveletScheme2}
    \Input{variables $\mathbf{x}$, list of neural networks $t$}
    Initialize $FactorOut$ as empty list\;
    \While{{\code{size(}$\mathbf{x}$\code{)} $\neq 2 \times 2$}}
    {
      Separate $\mathbf{x}$ into $\textbf{L}$, $\textbf{H}^a$, $\textbf{H}^b$ and $\textbf{H}^c$ as in Fig.~\ref{fig:partition}(b)\;
      Initialize counter $k$ as $0$\;
      \While{$k$ $<$ \code{len(}$t$\code{)}}
      {
        \Switch{$k$ mod $4$}
        {
          \Case{$0$}
          {
            $\textbf{L} = \textbf{L} + t[k]([\textbf{H}^a, \textbf{H}^b, \textbf{H}^c])$\;
          }
          \Case{$1$}
          {
            $\textbf{H}^a = \textbf{H}^a + t[k]([\textbf{L}, \textbf{H}^b, \textbf{H}^c])$\;
          }
          \Case{$2$}
          {
            $\textbf{H}^b = \textbf{H}^b + t[k]([\textbf{L}, \textbf{H}^a, \textbf{H}^c])$\;
          }
          \Case{$3$}
          {
            $\textbf{H}^c = \textbf{H}^c + t[k]([\textbf{L}, \textbf{H}^a, \textbf{H}^b])$\;
          }
        }
        $k$ ++\;
       }
       Set $\mathbf{x}$ equals $\textbf{L}$\;
       Append $FactorOut$ with $[\textbf{H}^a, \textbf{H}^b, \textbf{H}^c]$\;
      }
      \Return{$\mathbf{x}$, $FactorOut$}
    \end{algorithm}

\begin{algorithm}[hb!]
  \SetKwInOut{Input}{input}
  \caption{2D learnable wavelet transformation (scheme 2, follows traditional 2D wavelet convention)}
  \label{alg:2dwaveletScheme1}
    \Input{variables $\mathbf{x}$, list of neural networks $t$}
    Initialize $FactorOut$ as empty list\;
    \While{{\code{size(}$\mathbf{x}$\code{)} $\neq 2 \times 2$}}
    {
      Initialize counter $i$ as $0$\;
      \While{$i < 2$}
      {
       Initialize counter $k$ as $0$\;
       \While{$k <$ \code{len(}$t$\code{)}}
       {
         Separate $\mathbf{x}$ row-wise into $\textbf{L}$ and $\textbf{H}^a$ as in Fig.~\ref{fig:partition}(a)\;
         \If{$k$ is even}
         {
           $\textbf{L} = \textbf{L} + t[k](\textbf{H}^a)$\;
         }
         \Else
         {
           $\textbf{H}^a = \textbf{H}^a + t[k](\textbf{L})$\;
         }
         $k$ ++\;
       }
       Set $\mathbf{x}$ equals $[\textbf{L}, \textbf{H}^a]$\;
       Transpose $\mathbf{x}$\;
       $i$ ++\;
      }
      Set $\mathbf{x}$ equals upper-left quarter of $\mathbf{x}$\;
      Append $FactorOut$ with upper-right, down-left, down-right quarters of $\mathbf{x}$\;
    }
    \Return{$\mathbf{x}$, $FactorOut$}
\end{algorithm}

\end{document}